\pdfoutput=1
\documentclass[10pt,emptycopyrightspace]{ewsn-proc}

\usepackage{graphicx}
\usepackage{balance}
\usepackage{comment}
\usepackage{times}    
\usepackage{url}      
\usepackage{nameref}
\usepackage{caption}
\usepackage{subcaption}
\usepackage{fixltx2e}
\usepackage{multirow}
\usepackage{pbox}
\usepackage{array}
\usepackage{color}

\usepackage[table]{xcolor}
\definecolor{redncs}{RGB}{196,2,51}
\definecolor{maroon}{cmyk}{0,0.87,0.68,0.32}

\newcolumntype{C}[1]{>{\centering\let\newline\\\arraybackslash\hspace{0pt}}m{#1}}
\newcolumntype{L}[1]{>{\raggedright\let\newline\\\arraybackslash\hspace{0pt}}m{#1}}
\newcommand{\eg}{\emph{e.g.}}

\numberofauthors{3}
\author{
 Bradley McDanel\\
 Harvard University\\
  mcdanel@fas.harvard.edu
 \and
 Surat Teerapittayanon\\
 Harvard University\\
 steerapi@seas.harvard.edu
 \and
 H.T. Kung\\
 Harvard University\\
 kung@harvard.edu
}

\title{Embedded Binarized Neural Networks}

\begin{document}

\maketitle

\begin{abstract}
We study embedded Binarized Neural Networks (eBNNs) with the aim of allowing current binarized neural networks (BNNs) in the literature to perform feedforward inference efficiently on small embedded devices. We focus on minimizing the required memory footprint, given that these devices often have memory as small as tens of kilobytes (KB). Beyond minimizing the memory required to store weights, as in a BNN, we show that it is essential to minimize the memory used for temporaries which hold intermediate results between layers in feedforward inference. To accomplish this, eBNN reorders the computation of inference while preserving the original BNN structure, and uses just a single floating-point temporary for the entire neural network. All intermediate results from a layer are stored as binary values, as opposed to floating-points used in current BNN implementations, leading to a 32x reduction in required temporary space. We provide empirical evidence that our proposed eBNN approach allows efficient inference (10s of ms) on devices with severely limited memory (10s of KB). For example, eBNN achieves 95\% accuracy on the MNIST dataset running on an Intel Curie with only 15 KB of usable memory with an inference runtime of under $50$ ms per sample. 
To ease the development of applications in embedded contexts, we make our source code available that allows users to train and discover eBNN models for a learning task at hand, which fit within the memory constraint of the target device.
\end{abstract}

%
%


\section{Introduction}
Deep Neural Networks (DNNs), which are neural networks (NNs) consisting of many layers, are state of the art machine learning models for a variety of applications including vision and speech tasks. Embedded devices are attractive targets for machine learning applications as they are often connected to sensors, such as cameras and microphones, that constantly gather information about the local environment. Currently, for sensor data to be utilized in such applications, end devices transmit captured data in a streaming fashion to the cloud, which then performs prediction using a model such as a DNN. Ideally, we would leverage DNNs to run on the end devices, by performing prediction directly on streaming sensor data and sending the prediction results, instead of the sensor data, to the cloud.  

By classifying sensor data immediately on the embedded devices, the amount of information that must be transmitted over the network can be greatly reduced. As the number of devices grows on the network, the ability to summarize information captured by each device, rather than offloading raw data from the embedded device, becomes increasingly important. For example, in an object detection application, it requires $3.072$ KB to transmit a 32x32 RGB image to the cloud but only a single byte to send summary information about which object among, say, 256 candidate objects is detected in the image (assuming that inference could be performed directly on the device). This reduction in communication can lead to significant reduction in network usage in the context of wireless sensor networks.

However, the majority of these devices are severely constrained in processing power, available memory, and battery power, making it challenging to run inference directly. Therefore, we are interested in new approaches which can allow DNN inference to be run efficiently on these resource-limited devices. Additionally, many embedded devices do not have external memory, meaning that the entire DNN must fit in the on-device SRAM. Thus, we need to address the problem of minimizing the memory footprint required for inference. 
To this end, we leverage recent results on Binarized Neural Networks (BNNs), where 1-bit weights are used instead of 32-bit floating-point weights~\cite{courbariaux2015binaryconnect}. BNNs thus realize a 32-fold gain in reducing the memory size of weights, making it possible to fit much larger DNNs on device. Unfortunately, while the weights in BNNs are binary, the temporary results stored between layers in inference are floating-points. Even for small BNNs (\eg,~a one-layer convolutional network), the temporaries required for inference are significantly larger than the binary weights themselves, making inference on device not possible.

In this paper, we propose embedded binarized neural networks (eBNNs), which achieve a similar 32x reduction to BNNs in the memory size of the intermediate results used during layer-by-layer feedforward inference. By reordering the computation of inference in a BNN, eBNN preserves the original structure and accuracy of the network, while significantly reducing the memory footprint required for intermediate results. With eBNN, we demonstrate that it is possible to quickly perform DNN inference (10s of ms) on embedded devices with 10s of KB of memory. In Section~\ref{sec:eval}, we demonstrate the performance of eBNN on multiple neural network architectures with two datasets.
Additionally, we make a second contribution addressing the challenge of programming embedded devices with severe resource constraints. To ease programming in these settings, we propose a cloud-based service, which automates model learning and eBNN code generation for embedded devices.

\section{Related Work and Background}
\label{sec:background}
In this section, we provide a short introduction on the components (referred to as layers) used in deep neural networks (DNNs) and show how these layers are extended to binarized neural networks (BNNs) used in our approach.

\subsection{Deep Neural Network Layers}
A DNN is a machine learning method composed of multiple layers. DNNs have seen tremendous success in recent years, across many different input modalities, due to their ability to learn better feature representations, leading to higher accuracy, as more layers are added to the network. We briefly describe the important layers and functions of DNN feedforward inference that we use in this paper.

\begin{itemize}
    \item A \textit{Fully Connected Layer} has full connections to all neurons in the previous layer and can be viewed as a matrix multiplication between the input and weights of the layer. For classification tasks, these layers are typically used at the end of a NN to map the output to a vector of length $n$ where $n$ is the number of classes.
    \item A \textit{Convolutional Layer} consists of a set of filters typically learned during model training via supervised back propagation, which can identify discriminative patterns found in the input sample. Each filter is convolved with the input from the previous layer. Filters can have one or more dimensions based on the target application. In this paper, which considers 2-dimensional objects, each filter has two dimensions.
    \item \textit{Max Pooling} aggregates the input from the previous layer using the \textit{max} operation. This is typically used after convolution in order to combine extracted features from local regions in the input signal.
    \item \textit{Batch Normalization} is a normalization technique used to address the covariate shift problem found in DNNs as demonstrated in~\cite{ioffe2015batch}, and is used to improve the training time and the accuracy of the network.
    \item \textit{Activation Function} are nonlinear functions which are applied to the output of a layer to improve the representative power of the network. In our paper, we use a binary activation function discussed in the next section.
\end{itemize}

\subsection{Binarized Neural Networks}
In 2015, Courbariaux et al. proposed BinaryConnect, a method of training DNNs where all propagations (both forward and backward step) use binary weights~\cite{courbariaux2015binaryconnect}. In 2016, they expanded on this work with BinaryNet and formally introduced Binarized Neural Networks (BNNs)~\cite{courbariaux2016binarynet}. The second paper provides implementation details on how to perform efficiently binary matrix multiplication, used in both fully connected and convolutional layers, through the use of bit operations (xnor and popcount). In BNNs, all weights of filters in a layer must be $-1$ or $1$ (which is stored as $0$ and $1$ respectively) instead of a 32-bit floating-point value. This representation leads to much more space efficient models compared to standard floating-point DNNs. A key to the success of BNNs it the binary activation function, which clamps all negatives inputs to $-1$ and all positive inputs to $1$.  XNor-Net provides a different network structure for BNNs where pooling occurs before binary activation~\cite{rastegari2016xnor}. In this paper, we use the BNN formulation described in the BinaryNet paper. 

These networks have been shown to achieve similar performance on several standard community datasets when compared to traditional deep networks that use float precision. Research on BNNs thus far has primarily focused on improving the classification performance of these binary network structures and reducing the training time of the networks on GPUs. While the 32x memory reduction from floats to bits of the weights makes BNNs an obvious candidate for low-power embedded systems, current BNN implementations are for large GPUs written in one of several popular GPU frameworks (Theano, Torch)~\cite{team2016theano,collobert2002torch}. However, the computational model of GPUs is organized for high parallelism by reusing large temporary buffers efficiently. This computational model is a poor fit for embedded devices that have no hardware-supported parallelism and has only a relatively small amount of memory. In Section~\ref{sec:model}, we show that the optimal order of computation changes drastically when transitioning from a GPU environment with large memory and high parallelism to an embedded environment with small memory and no parallelism. Our implementation optimizations based on computation reordering are general and can be applied to other BNN structures.

\section{Embedded Binarized Neural Networks}
\label{sec:model}
In this section, we introduce embedded Binarized Neural Networks (eBNNs), which reorganize the order of computation in standard BNNs by combining multiple layers into fused blocks. Each fused block uses binary rather than floating-point temporaries to hold the intermediate results between the layers. Binarization of temporaries is significant, because as we show in Section~\ref{sec:eval}, floating-point temporaries consume a large portion of the total memory of embedded devices, in some cases making it impossible to run BNNs on embedded devices at all. For example, the Intel Curie, used in our evaluation, has 15 KB of usable SRAM memory. The smallest possible 1-layer convolutional BNN (a single 3x3 convolutional filter with 3 channels) would require only $22$ B to hold the neural network parameters (4 floating-point for Batch Normalization parameters and 2 bytes for each channel of the single convolutional filter), but $8.11$ KB to hold the floating-point temporaries (26 x 26 x 3 floating-points) for a 28 x 28 RGB input image. These temporaries make it difficult to realize the advantages of binarized representation of a BNN in the context of embedded devices.

\subsection{Fused Binary Blocks}
\begin{figure}[b!]
  \centering
    \includegraphics[width=\linewidth]{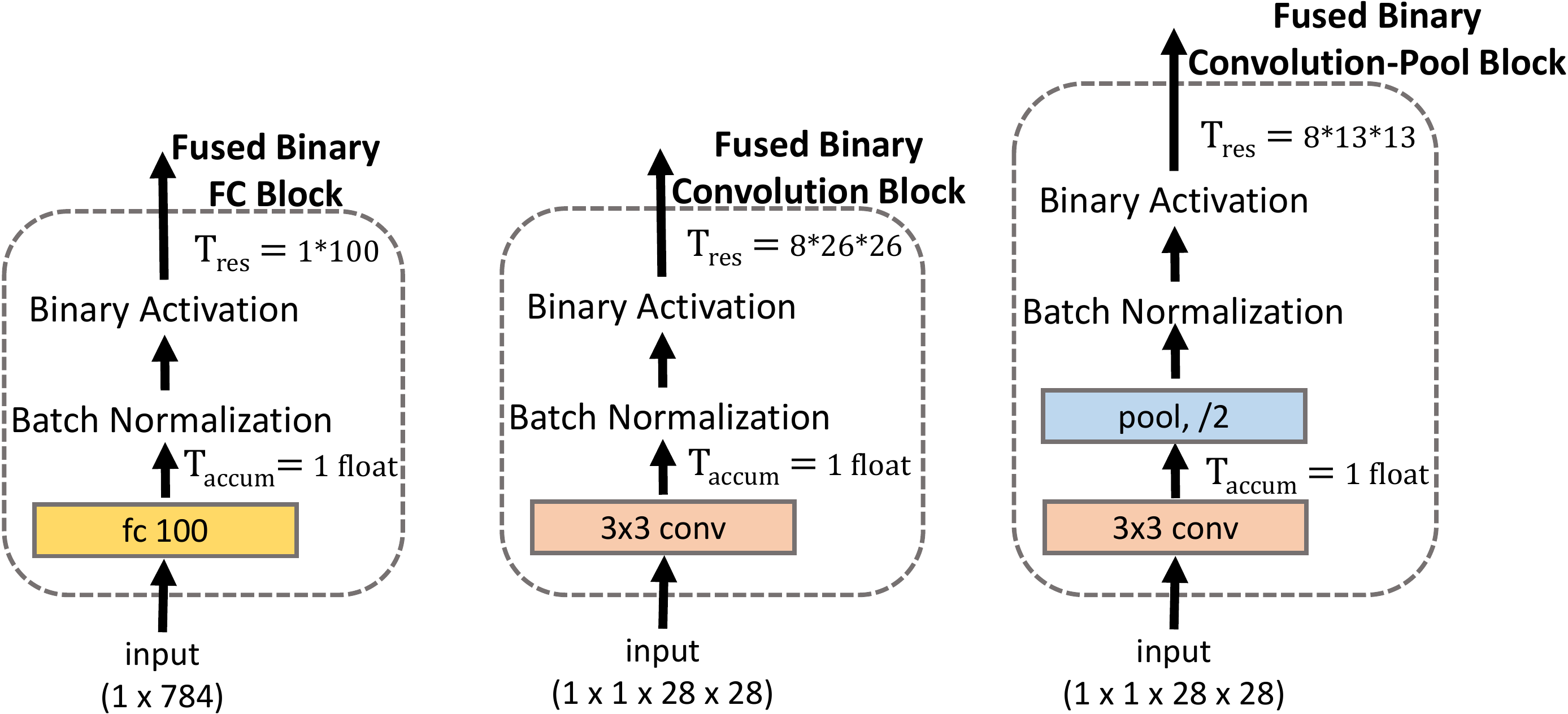}
  \caption{Structure of fused binary blocks, where $T_{accum}$ = 1 float means that temporary $T_{accum}$ is a single 32-floating-point number, that holds accumulated floating-point values used in all layers in the block. $T_{res}$ stores the binary output of the block, for 8 filters in this case, which is used as input by the next block in the network. These blocks are presented in the context of 2D input data, but can also be used with 1D and 3D inputs.}
  \label{fig:binary_block}
\end{figure}

We adopt the concept of NN blocks used in several deep network architectures~\cite{he2016identity}. Network blocks each contain one or more NN layers that can be stacked in a modular fashion just like standard NN layers. In this paper, we introduce fused binary blocks, which reorder the computation of multiple NN layers within a block in order to minimize the space required for intermediate temporary results. Figure~\ref{fig:binary_block} shows three fused binary blocks we use in this paper, operating on a 28 x 28 input sample: a Fused Binary Fully Connected (FC) Block, a Fused Binary Convolution Block and a Fused Binary Convolution-Pool Block. Internally, each block rearranges the order of computation to use only a single floating-point accumulator, $T_{accum}$, and outputs binary rather than floating-point results, $T_{res}$, which are then used as input to the next block.

\begin{figure}[b!]
  \centering
    \includegraphics[width=1\linewidth]{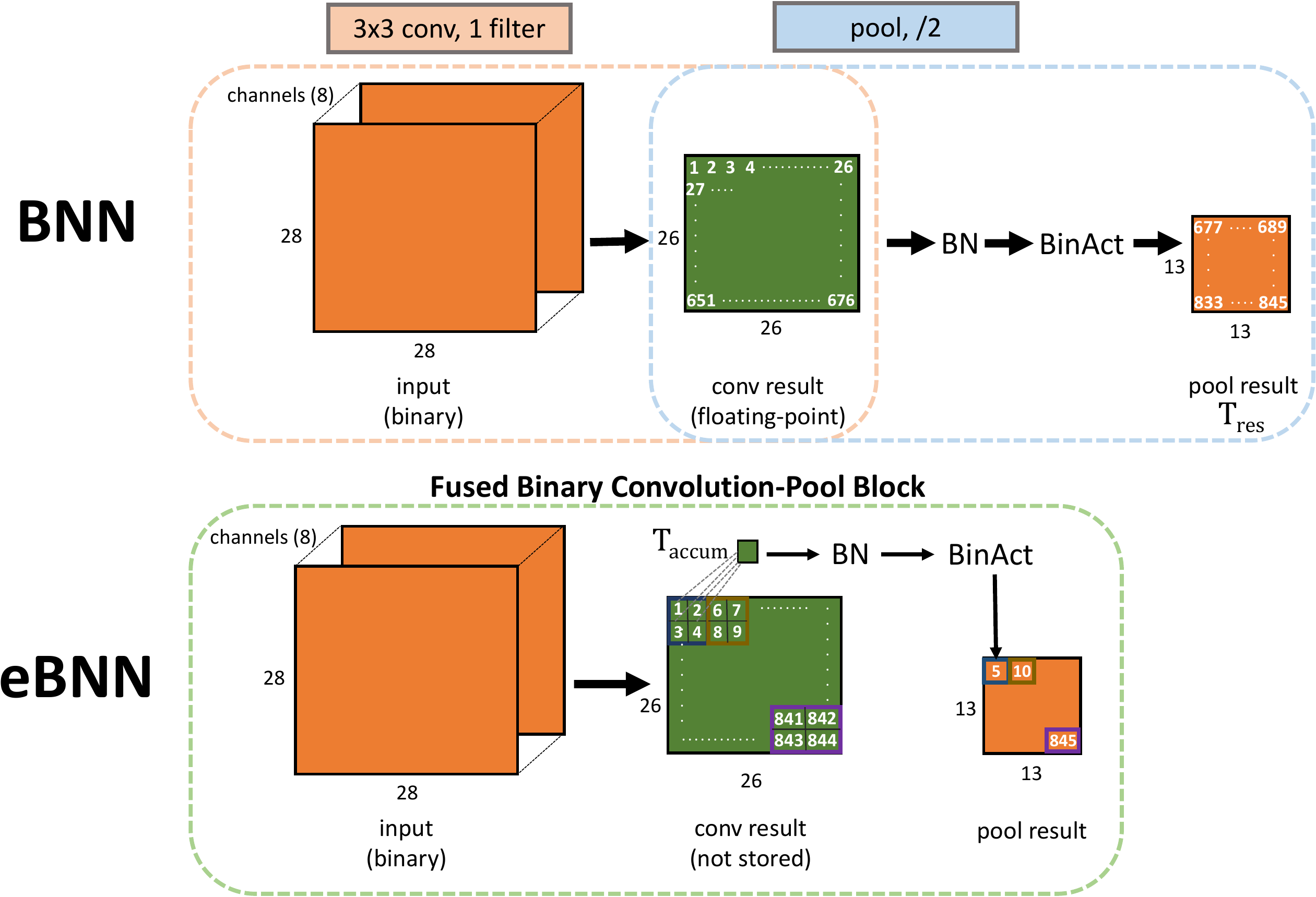}
  \caption{Comparison of temporary memory usage for BNN and eBNN. BN is Batch Normalization. BinAct is the Binary Activation function. For BNN, in the convolution phase, convolution is performed on each 3x3 patch, and the result is stored (in green) as input for the pooling layer. By comparison, eBNN uses Fused Binary Convolution-Pool Blocks which fuse the convolution and pooling layer and require one accumulator, $T_{accum}$, to store the intermediate convolution result. In the case of max pooling, $T_{accum}$ is computed by serially comparing each convolution result in a region against the current max. For a given input area, once all convolution results are accumulated into $T_{accum}$ the max-pooled result is sent through the BN and BinAct functions, which quantizes the floating-point to a single binary value, before being stored in $T_{res}$. The order of the convolution and pooling operations are shown from 1 to 845.}
  \label{fig:eBNN_comparison}
\end{figure}

Figure~\ref{fig:eBNN_comparison} shows the order of computation of a binary convolution layer followed by a pooling layer, as in BNN, compared to the fused binary convolution-pool block, as in eBNN. In BNN, all the convolution results are first computed and stored as floating-point temporaries (the green block). After convolution is complete, pooling takes place on the output matrix from convolution. By doing the convolution and pooling layers in two phases, the network must store the output of the convolution layer in the floating-point representation requiring 26x26 float-point storage of temporaries.

In contrast, under eBNN, in the Binary Convolution-Pool block, the convolution and pooling operations are fused so that only a single convolution result is stored (in floating-point format) at a time. In the case of max pooling, only the maximum of the local pool window is stored. Once the entire pooling region has been accumulated into $T_{accum}$, the maximum passes through through batch normalization and a binary activation function and is stored as a single binary value in the $T_{res}$ result matrix. In the case of overlapped pooling, we recompute the values in the convolution result matrix which are used multiple times, in order to keep the lowest memory footprint possible. Figure~\ref{fig:eBNN_comparison} shows the computation required for a single result in $T_{res}$ with a single filter. In the general case of multiple filters, the process is repeated for each filter, and the output $T_{res}$ will have dimensions of filters x width x height. $T_{accum}$ is reused for each filter, as they are computed one at a time. This reordering prioritizes low memory usage over parallelism and is tailored specifically for embedded devices with small memory capacity.

\subsection{Memory Cost for Feedforward Inference}
In order to generate BNNs that fit within the memory constraints of the embedded devices, an accurate memory cost model for inference using a given network is required. We can break down the memory requirements of inference into two parts: the weight and normalization parameters of the BNN and the temporaries required to hold intermediate results between network layers. Since \textbf{\textit{eBNN does not modify the original BNN network}}, but instead only reorders computation, the size of the model parameters are identical for both networks.

For BNNs, the memory size of the temporaries is the output dimensionality of the largest NN layer in floating point format. Even though the eventual result will be stored in binary, the intermediate results are stored in floats so that operations in subsequent layers (such as pooling and batch normalization) can be readily parallelized over parallel GPU cores. In eBNN, the memory size is also proportional to the output dimensions of the largest NN layer, but in binary rather than floating-point format, as only binary values are stored. This leads to a theoretical 32x savings in memory for temporary values. In practice, we store each row of output on byte boundaries for performance reasons, so there can be a small amount of memory waste. At worst, this waste is 7 bits per row of $T_{res}$. However, eBNNs can be designed, through the convolution and pooling stride parameters, so that the resulting $T_{res}$ rows are always divisible by $8$ and therefore have no waste.

Formally, we write the required memory cost of inference for eBNN in bits $\textbf{M}$ as $\textbf{M} = \textbf{P} + 2\textbf{T}$ where $\textbf{P} = \sum_{i=1}^N p_i$ represents all of the parameters, where $p_i$ denotes the weights and Batch Normalization parameters for layer-i of an N-layer network, and $\textbf{T} = \max_{i=1..N} T_{res}^i$ is the required temporaries for the widest layer in the network. We need twice the maximum layer in temporaries to hold both the binary input and output.


\section{Evaluation}
\label{sec:eval}
In this section, we evaluate the proposed eBNN on an image classification task; however, the same concept could apply to other types of common tasks in wireless sensor networks (WSN) such as sensors detecting intrusion, landslide and forest fire or activity recognition using body-area WSN. Specifically, several different NN structures of varying depths and two standard community datasets (MNIST, CIFAR10)~\cite{lecun1998mnist,krizhevsky2009learning} are used to evaluated the proposed eBNN. The device used in our evaluation is the Arduino 101 based on Intel Curie (32MHz 32-bit Quark SoC) with 24 KB of SRAM.

Given the small nature of the network we can fit inside the Curie, we are interested in evaluating the networks on data samples that are relatively easy to classify. In many settings, embedded devices will work in a fixed or constrained setting, with a smaller space of possible data samples. For CIFAR10, we created a simpler evaluation dataset (CIFAR10-Easy), by training a large network (a standard DNN with 5 convolutional layers), sorting test samples based on the entropy at the final softmax layer (before classification), and taking the 10\% of samples with lowest entropy.  In Section~\ref{sec:service}, we introduce a service model to train eBNN networks for a target dataset. We envision this may be used to build a model to classify a specific task using personalized dataset.

The same computation reordering principle described in Section~\ref{sec:model} is applied to all of the networks. Our first network structure is a Multilayer Perceptron (MLP) composed of fused binary FC blocks. We use a 1 hidden layer (MLP-1) and 2 hidden layers (MLP-2) networks for our evaluation. We are interested in these shallow MLPs due to their wide applicability and fast inference runtime. Additionally, the fused binary FC block is needed for the last layer of eBNN convolution networks, so we are interested in evaluating the performance of the block in isolation.

\begin{figure}
  \begin{minipage}[ht!]{0.25\linewidth}
    \includegraphics[width=\linewidth]{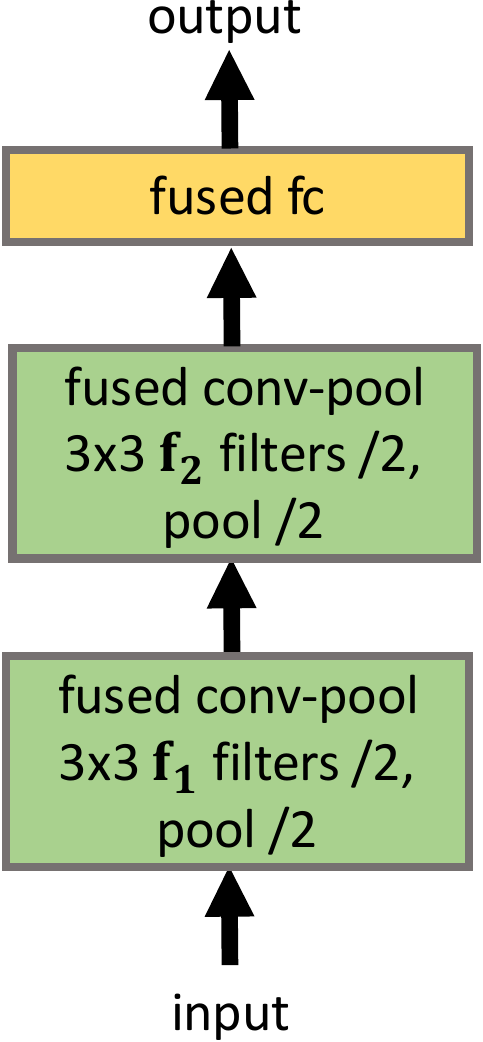}
  \end{minipage}\hfill
  \begin{minipage}[c]{0.7\linewidth}
    \caption{An eBNN used in evaluation consisting of two layers of fused binary convolution-pool blocks. \textbf{f\textsubscript{1}} and \textbf{f\textsubscript{2}} represent the number of filters in the corresponding layer. The first $/2$ in the layer description is the stride for convolution, and the second pool $/2$ is the stride for pooling.} \label{fig:fused_network}
  \end{minipage}
\end{figure}

We also evaluate eBNN applied to convolutional neural networks. Figure~\ref{fig:fused_network} shows an eBNN fused binary convolutional-pool network that is used in our evaluation. The network is a two layer network consisting of fused binary convolution-pool blocks. Binary convolutional networks have more concise representations in terms of model parameters, but much larger required temporary storage to hold the intermediate results between layers. These networks show off the importance of reordering the computation as described in the previous section. As mentioned earlier, even a 1-layer convolutional network with a single filter does not fit within the memory constraints (15KB) of the device (Intel Curie) using the BNN approach. For the sake of completeness, we also evaluate convolutional networks with larger strides in convolution (stride = 3), but without pooling steps. Practically, these convolution-only networks may still achieve reasonable accuracy but run faster due to the absence of the pooling step.

Additionally, for the MNIST dataset, we also consider two low-energy NN models: Conv-1-LE-I and Conv-1-LE-II. These models use less memory and achieve reasonably good performance despite the smaller model size. We use these models to show that eBNN can be tuned to suit particular memory, energy and accuracy requirements of a system.

We limit each eBNN to 15 KB as a maximum cutoff point to ensure each model fits within the device. Some of the available memory is needed to store program code and handle the incoming input sample so not all 24 KB is used for the model. Input samples are processed in a streaming fashion (batch size = 1) in order to model a setting where data would be continuously acquired through attached sensors.

\begin{figure*}[h]
\begin{minipage}[t]{.64\textwidth}
\includegraphics[width=.52\textwidth]{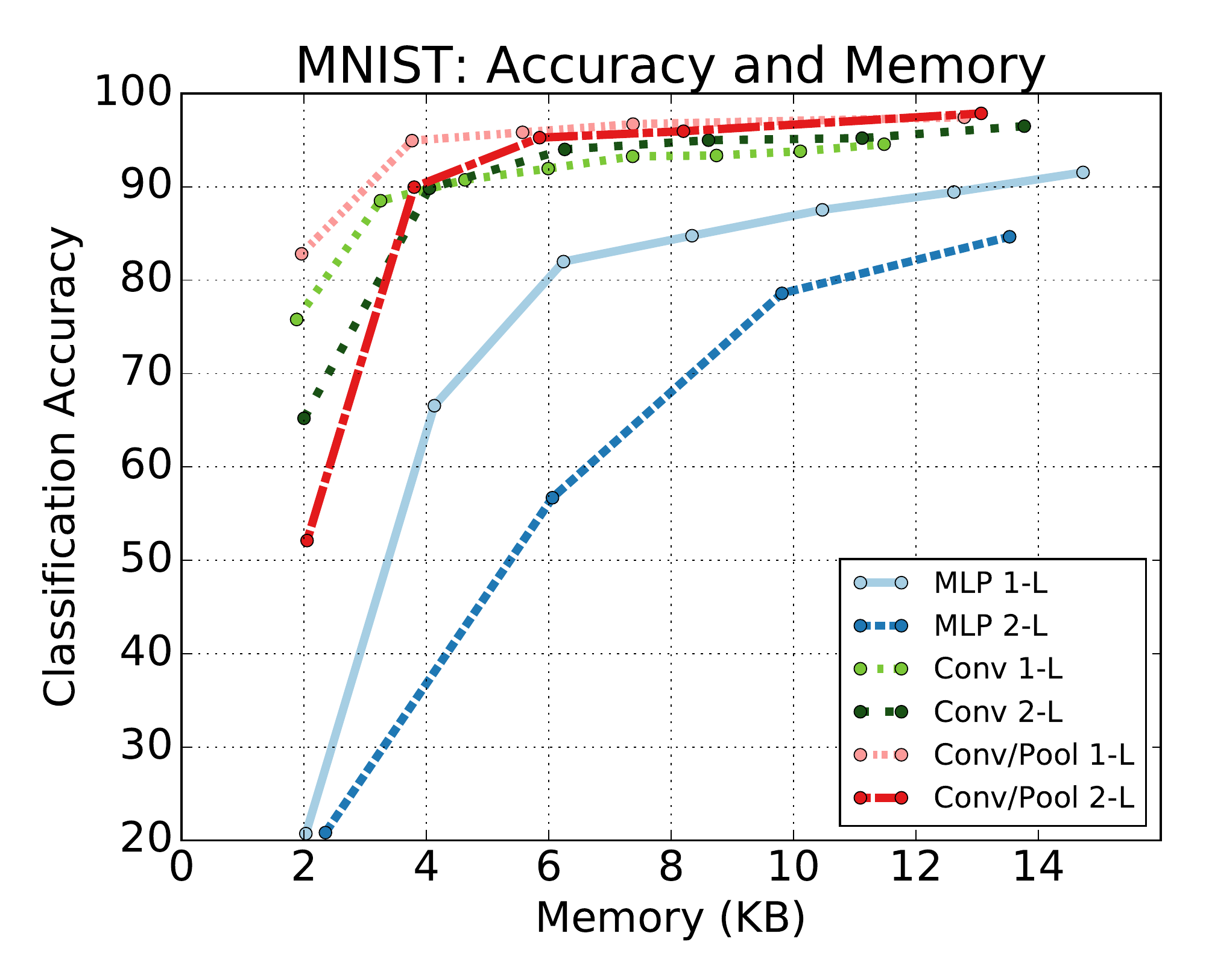}\hfill
\includegraphics[width=.52\textwidth]{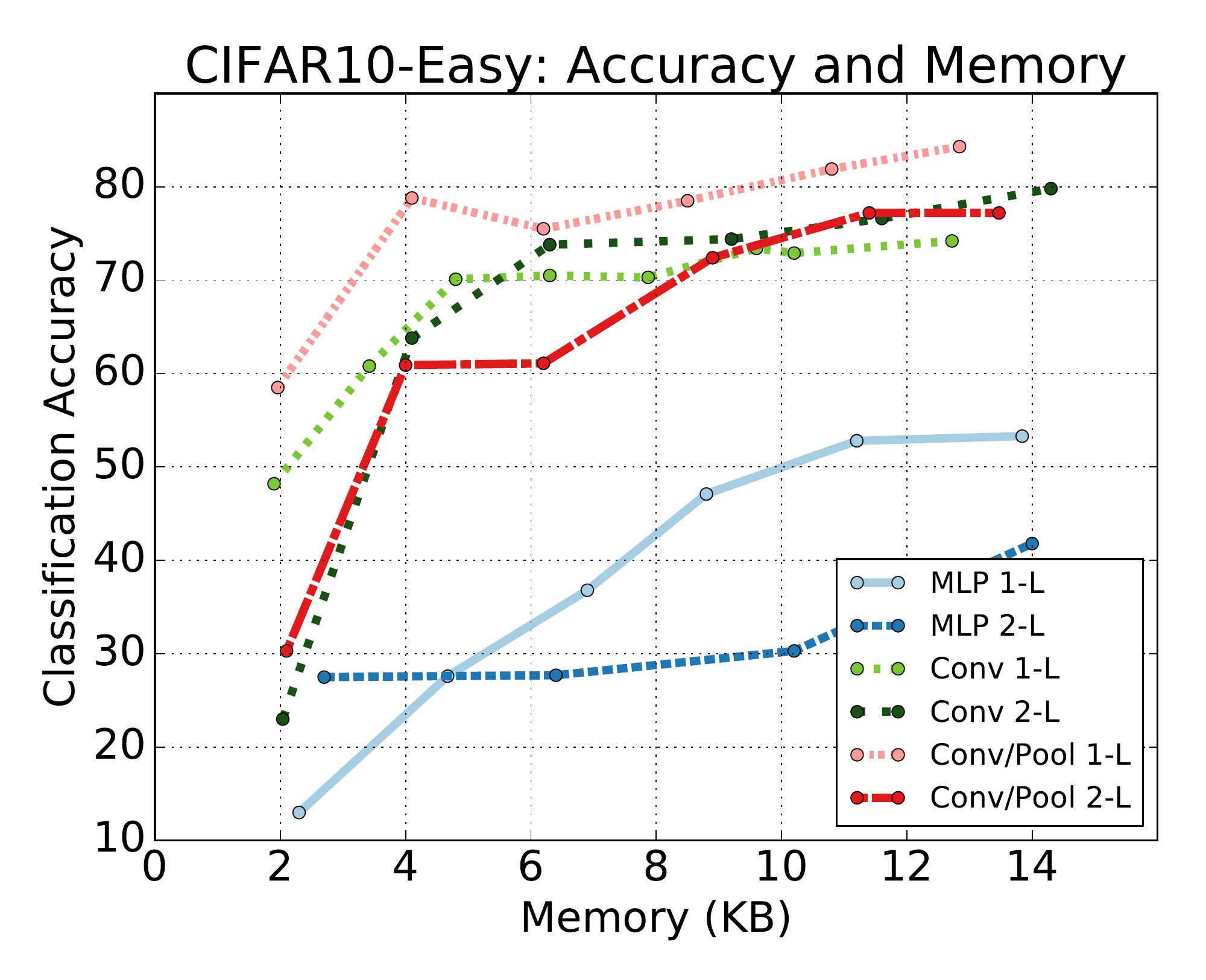}

\caption{The accuracy and memory size of the eBNN evaluation models for the MNIST and CIFAR10-Easy datasets, as the number of model parameteres increases. The accuracy shown at each point for eBNN is the same as that for the corresponding BNN model. For all models, the memory size of temporaries comprises at most $3\%$ of the overall memory.}
\label{fig:accuracy_memory}
\end{minipage}
\hfill
\begin{minipage}[t]{.34\textwidth}
\includegraphics[width=\textwidth]{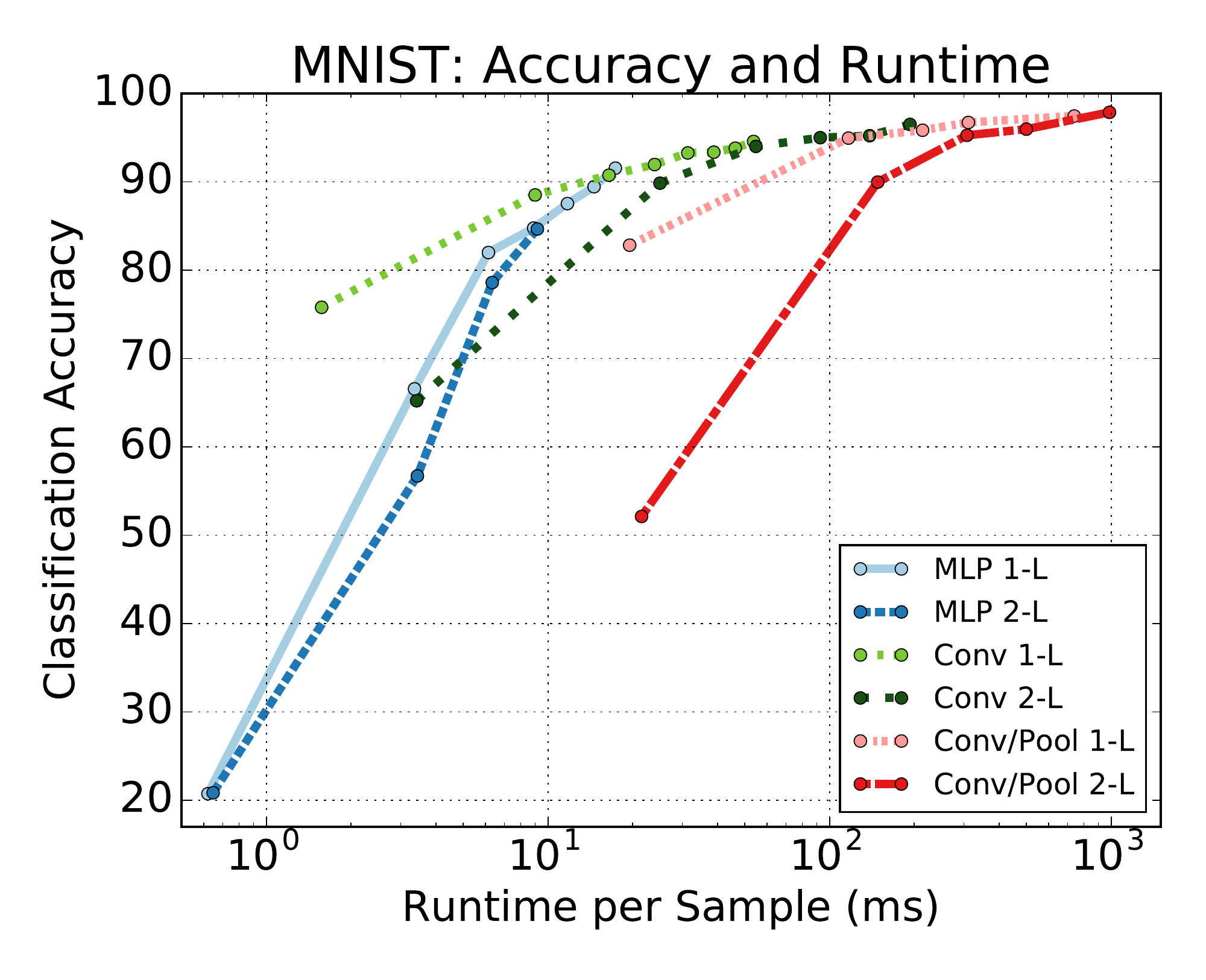}
\caption{The runtime per sample of each network configuration as number of network parameters are increased evaluated on MNIST. The x-axis is a log scale.}
  \label{fig:runtime}
\end{minipage}
\end{figure*}



Figure~\ref{fig:accuracy_memory} shows the accuracy and runtime of each eBNN structure on the MNIST and CIFAR10-Easy datasets. The main observation is that with eBNN the temporary overhead is only a small portion of the entire memory required for inference across all NN structures. For all networks, the temporaries take up at most $3\%$ of the entire memory required for inference. The network parameters (i.e., filters and batch normalization parameters) take up the remaining $97\%$ of device memory. This low memory footprint allows for substantially larger networks to fit on the device than would otherwise be possible and therefore enables a much higher classification accuracy.  

Table 1 provides detailed results for each eBNN network corresponding to last point in curve for each network in Figure~\ref{fig:accuracy_memory}. For MNIST, the 1-layer MLP (MLP-1) network achieves an accuracy of $91.54\%$ for the best MLP model within the memory limit. The 2-layer MLP (MLP-2) performs worse than the single layer version. The 1-layer MLP also outperforms the 2-layer MLP on the CIFAR10-Easy dataset. This suggests that in constrained settings, more neurons in a single layer may provide a more substantial accuracy gain than an additional layer where the neurons are split between layers. The 2-layer Convolution-Pool network provides the best accuracy for MNIST, reaching $97.86\%$. The 1-layer and 2-layer Convolution only networks (Conv-1 and Conv-2 respectively) achieve slightly reduced accuracy compared to the pooling variation (ConvPool-1 and ConvPool-2). For larger networks, with additional network layers, pooling has been shown to be an important step~\cite{scherer2010evaluation}, and even in a two layer network, we see pooling improves performance.

Figure~\ref{fig:runtime} shows the average runtime, in wall clock time, of the various models processing a single MNIST sample on the device. All network settings shown in Figure~\ref{fig:accuracy_memory} run inference under 1 second, with smaller networks running in the 10s of ms. Interestingly, we see that the convolution model without pooling provides the best trade-off between runtime and accuracy. The convolution-pool networks provide the best accuracy, but at the cost of some additional runtime overhead caused by the pooling operations. The MLP models perform similarly in terms of runtime to the convolution only networks but with reduced accuracy.

In order to understand the communication and energy costs of eBNN, we define a couple terms used in comparison to an approach that offloads all sensor data to the gateway/cloud. Communication Reduction (CR) is the size of the sensor input sample over the size of the classification label. Energy Gain (EG) is the gain in energy by running eBNN locally on the device versus the amount of power it takes to transmit (using BLE, Wi-Fi, etc.) the input sample to another device (which then performs classification). The Arduino 101 uses 0.150 mW while idle,  0.250 mW during computation, and 0.200 mW while transmitting over BLE. 
For MNIST, it takes 24.5 ms to transmit the 784 bytes input (28x28 grayscale image) with energy of 6.125 mW. By comparison, it takes 31.25 $\mu$s to transmit the 1 byte classification label with energy of 7.8125 $\mu$W. This leads to a CR of 784x in this case. The energy of eBNN is determined by the runtime of inference for a given model. We see in Table~\ref{table:results} for MNIST, that the selected MLP-1, MLP-2, Conv-1-LE-I and Conv-1-LE-II models use less energy than is required to transmit the sensor input. Specifically, Conv-1-LE-II achieves an EG of 1.5x with maintaining an accuracy of 91\%. For the larger models, eBNN still achieves the same CR while improving the classification accuracy, for an increase in energy. Similar trends are seen for the CIFAR10-Easy dataset. The energy gain of eBNN will increase when other wireless protocols such as Wi-Fi and 4G are used which can transmit farther than BLE but consume more energy.

In this paper, we only consider the performance of eBNN from the viewpoint of a single embedded device. However, in a network environment with many devices, the high CR shown in eBNN is even more important. For instance, in wireless sensor networks connecting these devices, transmitting less amount of data reduces congestion, leading to reduced transmission time and energy consumed. Additionally, in the case of multi-hop networks, the total energy required to transmit the input data from a sensor node to a gateway is multiplied by the number of hops it takes to reach the gateway. Therefore, the CR of eBNN provides larger energy savings in multi-hop scenarios. 

\begin{table}[htp]
\centering
\small
\caption{Selected eBNN performance results. 
}
\label{table:results}
\begin{tabular}{ lllll }
\hline
\rowcolor{gray!20}{\textbf{MNIST}} \span \span \span \span \\
\hline
\textbf{Model}                     & \textbf{Acc.} (\%) & \textbf{Time} (ms)    & \textbf{Mem.} (KB) &  \textbf{Enrg.} (mWs)       \\\hline
\rowcolor{maroon!20} MLP-1         & 91.54              & 17.35                 & 14.73              &  5.37                        \\
\rowcolor{maroon!20} MLP-2         & 84.65              & 9.17                  & 13.53              &  4.95                        \\
\rowcolor{maroon!5}  Conv-1        & 94.56              & 53.72                 & 11.48              &  19.96                        \\
\rowcolor{maroon!5}  Conv-2        & 96.49              & 193.02                & 13.77              &  63.02                        \\
\rowcolor{maroon!20} ConvPool-1    & 97.44              & 739.34                & 12.79              &  213.63                        \\
\rowcolor{maroon!20} ConvPool-2    & 97.86              & 886.53                & 13.07              &  243.98                        \\
\rowcolor{magenta!10}  Conv-1-LE-I    & 91.95              & 23.91                  & 5.99               &  6.02                        \\
\rowcolor{magenta!10}  Conv-1-LE-II   & 90.74              & 16.47                 & 4.63               &  4.15                        \\\hline
\rowcolor{gray!20}{\textbf{CIFAR10}\textbf{-Easy}} \span \span \span \span \\
\hline
\textbf{Model}                     & \textbf{Acc.} (\%) & \textbf{Time} (ms)    & \textbf{Mem.} (KB) &  \textbf{Enrg.} (mWs)  \\\hline
\rowcolor{maroon!20} MLP-1         & 52.30              & 21.29                 & 13.84              &  4.37                        \\
\rowcolor{maroon!20} MLP-2         & 41.80              & 19.65                 & 14.00              &  2.31                        \\
\rowcolor{maroon!5}  Conv-1        & 74.20              & 79.21                 & 12.72              &  13.54                        \\
\rowcolor{maroon!5}  Conv-2        & 79.80              & 250.08                & 14.30              &  48.64                        \\
\rowcolor{maroon!20} ConvPool-1    & 84.30              & 847.72                & 12.84              &  186.31                        \\
\rowcolor{maroon!20} ConvPool-2    & 77.20              & 968.18                & 13.47              &  223.41                        \\\hline
\end{tabular}
\end{table}

\section{Implementation}
\label{sec:service}
Our eBNN feedforward inference implementation is written in C and has a corresponding Python version in Chainer~\cite{chainer_learningsys2015} that is utilized to train the BNN models. Each fused binary block has a Chainer link with a modification that enables it to output the model parameters for that layer into a generated C file. Once trained in Python, the network is automatically converted into a standalone C header file with corresponding inference code. We validated the correctness of the C implementation by comparing the output at each stage to the Python version. Our codebase is open source and is available here: \url{https://gitlab.com/htkung/ddnn}.

In addition to the software, we implemented a service model which allows users to train and discover eBNNs that provide the best prediction accuracy while fitting within the memory constraints of a specified device. Python code to screen over this search space is also provided. Since the training process can use GPU implementations of BNN, we can quickly run an optimization (on the order of minutes) that explores various deep neural network models with different parameters to find the best setting for a particular device and task. We envision this as a useful rapid prototyping tool to run deep networks on embedded devices. Practically, it is challenging to port a BNN network written in another language into an embedded C environment. Our codebase aims to alleviate this issue. 

\section{Conclusion}
\label{sec:conclusion}
BNNs have huge potential for embedded devices, due to the 32x reduction in model weights, which can allow deep networks to fit within the device. However, the temporaries required to hold intermediate results between layers represents a substantial portion of memory required to run inference. Temporaries are especially significant for convolutional networks; for BNN the temporary overhead is larger than the memory size (15KB) of our experimental device even for a single filter 1-layer convolutional network. Our proposed eBNN achieves a 32x space reduction for temporaries. The eBNN scheme is based on reordering of computation (including recomputation for overlapping pooling if required) without changing the structure of original BNNs, thus preserving the same accuracy as the original network. The optimizations used by eBNN to mitigate the overhead of temporaries is a fundamentally different from that in GPU programming for DNN, that is, our goal is minimizing memory, rather than using large memory in order to allow parallel processing. Additionally, we proposed a service model to ease programming on small devices, and to automatically carry out trade-off screening between accuracy and memory requirement, as described in in Section~\ref{sec:eval}.

In this paper, we have demonstrated encouraging results on implementing DNN on embedded devices: run time can be in the 10s of ms on devices with memory as small as 10s of KB while achieving respectful recognition accuracy. The fact that a tiny wearable can do fast DNN inference is somewhat surprising. We are not aware of previous work that shows this was feasible. Future work includes implementing eBNN on FPGAs, which would be orders of magnitude faster and more power efficient than eBNN on the Curie due to being able to better exploit the binary structure of the network. Perhaps eBNN on FPGAs could even be faster than GPU implementation for the same power consumption. Compiler development for automatic generation of re-ordered computation for eBNN code could also be a fruitful direction for future work.

\section*{Acknowledgment}
  This work is supported in part by gifts from the Intel Corporation
  and in part by the Naval Supply Systems Command award under the
  Naval Postgraduate School Agreements No. N00244-15-0050
  and No. N00244-16-1-0018.


%
%
%
%
%

\bibliographystyle{abbrv}
\bibliography{references}
\end{document}